# Computer Vision Navigation for an Autonomous Vehicle Model Using Reinforcement Learning

Dinis Fernandes, Tiago Ribeiro, António Ribeiro, Carolina Lopes, Gil Lopes, A.Fernando Ribeiro
Lab. Automação e Robótica
University of Minho
Guimarães, Portugal
a86021@alunos.uminho.pt

## Abstract

This work aims to assess the capabilities of the reinforcement learning algorithm, Proximal Policy Optimization (PPO), in developing a policy capable of controlling the speed and steering angle of a vehicle model to navigate within simulated road lines.

The control system's input is a camera, and its outputs are the rotational speed of the rear wheels and the steering angle of the front wheels, controlling the vehicle's speed and direction, respectively. The algorithm is trained and tested in a simulated environment, where it receives image frames from the vehicle's camera, processes them through a feature extraction algorithm, and feeds them into the policy. For each frame, the policy outputs speed and direction commands for the vehicle. The policy is iteratively updated to enhance its performance.
The reinforcement learning algorithm successfully generates a policy capable of navigating the vehicle in a simplified environment. Potential improvements are suggested to enhance the system's robustness in more complex scenarios.

Creating a simulation environment suitable for testing an advanced reinforcement learning algorithm is a complex task. This project offers a framework for other developers to test different approaches and algorithms. The tested algorithm, PPO, uses proximity intervals—an innovative approach in reinforcement learning. In some scenarios, the algorithm achieved better performance than predecessor algorithms, which are typically harder to implement and require tuning more hyperparameters.

## 1. Introduction

The concept of reinforcement learning (RL) is inspired by the way animals learn to perform tasks by taking actions and receiving rewards or punishments based on the outcomes of those actions. To choose an appropriate action, the agent must understand its state relative to the environment. The policy of the agent is the mechanism that maps states to actions. Based on the rewards or punishments received, the agent iteratively adjusts its policy to improve performance.

Modern RL algorithms typically employ neural networks as policies and evaluate policy performance using a loss function. The loss function quantifies the discrepancy between the desired and actual outcomes, with higher values indicating poorer performance. Consequently, the goal is to minimize this loss function. Gradients, a fundamental concept in calculus, are used to find the parameters that minimize the loss function, thereby refining the policy.

Policies trained with RL have demonstrated remarkable success across a variety of domains, often surpassing human performance. One of the most prominent examples is AlphaGo (Silver et al., 2016), an algorithm that defeated the world's best human Go players. Go is an exceptionally complex game with more possible board configurations than there are atoms in the universe, making it infeasible to tackle using traditional algorithms. Similarly, autonomous driving requires learning-based approaches rather than hard-coded solutions, as the aim is to develop an algorithm that learns to perform the task rather than explicitly solving it.

The Laboratory of Automation and Robotics (LAR) at the University of Minho has been highly active in autonomous driving research. Engineers at LAR developed an autonomous vehicle model that has been used in several projects to enhance its sensing and control capabilities. For example, one project aimed to create a model capable of reading road signs using YOLOv3 (Marques et al., 2022). Another project combined YOLOv3 with the Deep Deterministic Policy Gradient (DDPG) algorithm to develop a system capable of detecting and classifying road signs and responding accordingly (Andrade et al., 2022). Additionally, a separate project focused on training a policy for controlling the vehicle to perform parallel, angular, and perpendicular parking (Sousa et al., 2022).

The objective of this project is to employ an RL algorithm to train a policy capable of navigating the Portuguese National Robotics Championship autonomous driving track. The control system receives input from a camera and outputs the steering angle of the front wheels and the speed of the rear wheels. The vehicle model used in this study (Figure 1) is based on the design developed by LAR at the University of Minho.

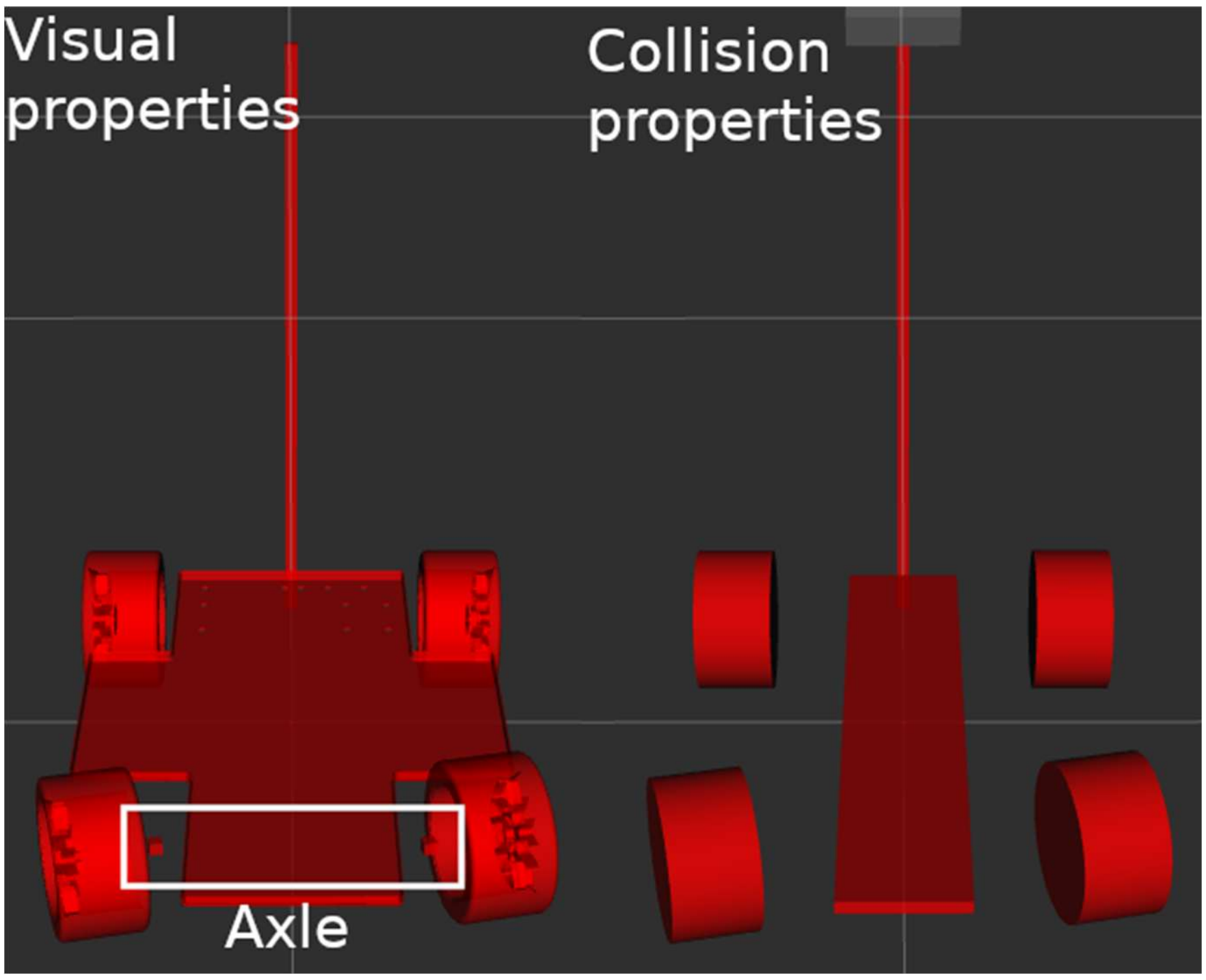

**Figure 1:** Simulation model of the autonomous vehicle developed by LAR

The first phase of the project focused on training the policy within a simulated environment, as real-world training is not an efficient process. Training in the real world would require continuous supervision of the algorithm's behaviour during the training period, posing risks to the hardware. The second phase would involve testing the trained policy on the physical vehicle model. However, this phase was not completed due to the time required to develop the simulation environment, implement the algorithm, and train the policy, which exceeded initial expectations. Nonetheless, the results obtained in the simulation suggest that this approach is feasible, with certain improvements outlined in the conclusion.

The simulation environment (Figure 2) was implemented using the Gazebo robotics physics simulator. The vehicle model was described and deployed in Gazebo, and the control of actuators and data acquisition from the camera were managed through the Robot Operating System (ROS). Feature extraction was implemented using OpenCV to process the frames captured by the camera. The implementation of the Proximal Policy Optimization (PPO) algorithm was carried out using TensorFlow 2.

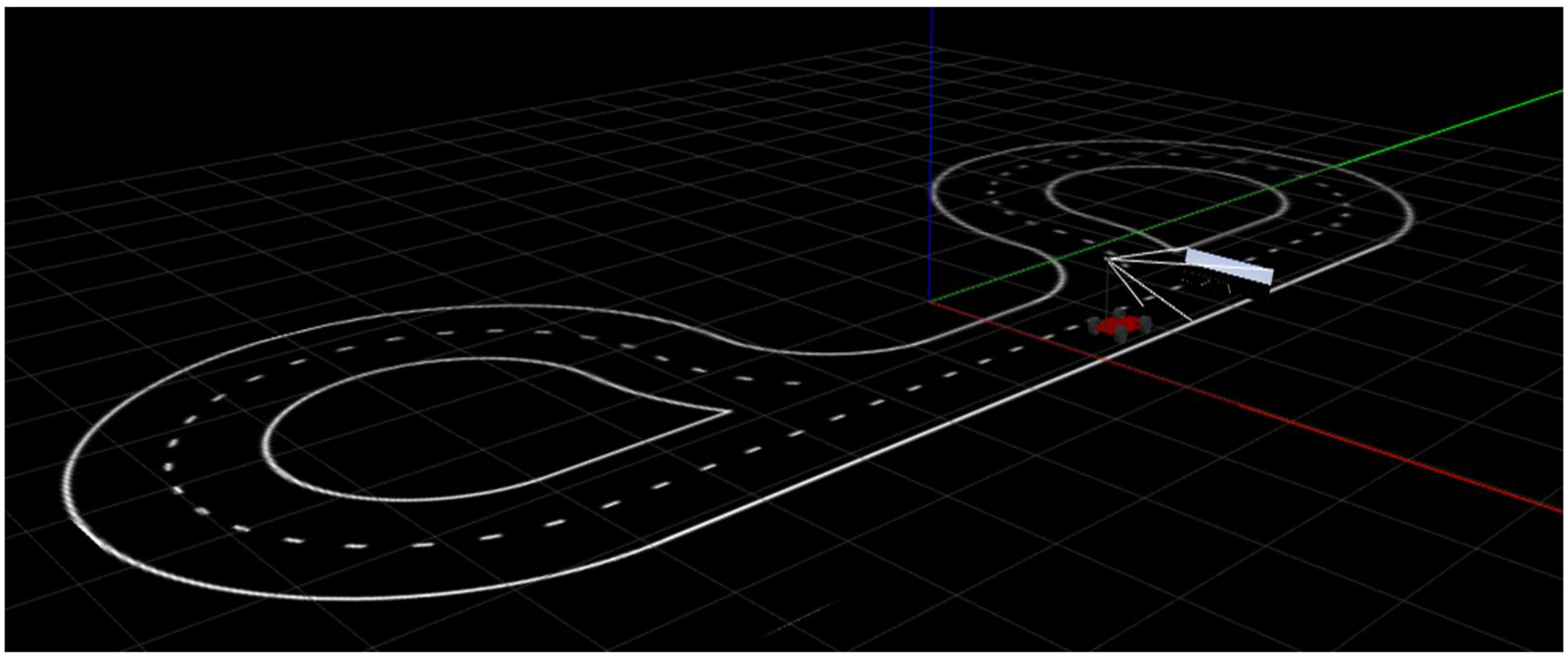

**Figure 2:** Portuguese robotics championship autonomous driving track in simulation with vehicle model

Additional resources, including the project code and demonstration video, can be found in the **Supplementary Materials** section.

## 2. Algorithm overview

The input to the algorithm is a camera frame that, after processing, becomes the state representation for the algorithm. The output consists of two values: one controlling the vehicle's direction and the other its speed. Specifically, the direction and speed are determined by selecting a value for the front wheels' axle angle and a value for the rear wheel rotation speed. These values are expressed in radians (rad) and radians per second (rad/s), respectively.

After training (Figure 3), the algorithm operates by acquiring frames from the camera. Each frame is processed by the feature extraction algorithm to extract relevant environmental data. The extracted features define the state, which is then fed into the neural network policy. The policy outputs the action values, which are used to control the wheels through the actuators.

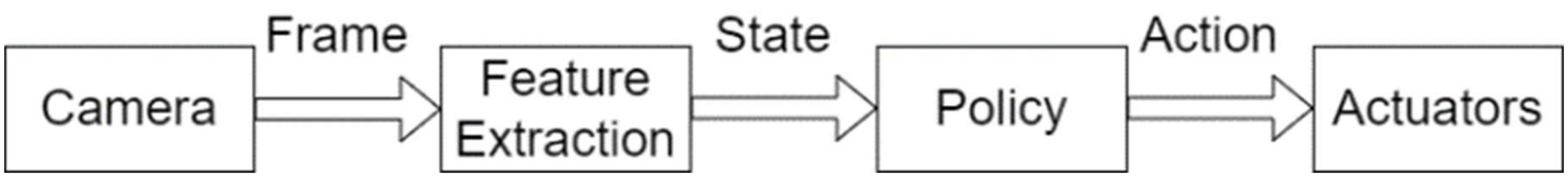

**Figure 3:** Overview of the algorithm after training the policy.

The training algorithm is repeated iteratively until the policy demonstrates satisfactory performance in navigating the track. It begins by retrieving states from the sensors, executing the corresponding actions through the actuators, and storing a tuple containing the necessary data for gradient updates. Once the buffer that stores these tuples reaches its capacity, the algorithm proceeds to the gradient update phase. During this process, the data stored in the buffer from the experience acquisition phase is used to update the neural network.

Figure 4 presents an overview of the training algorithm's workflow. Although the final training implementation includes some modifications, the overall process can be understood from this flowchart.

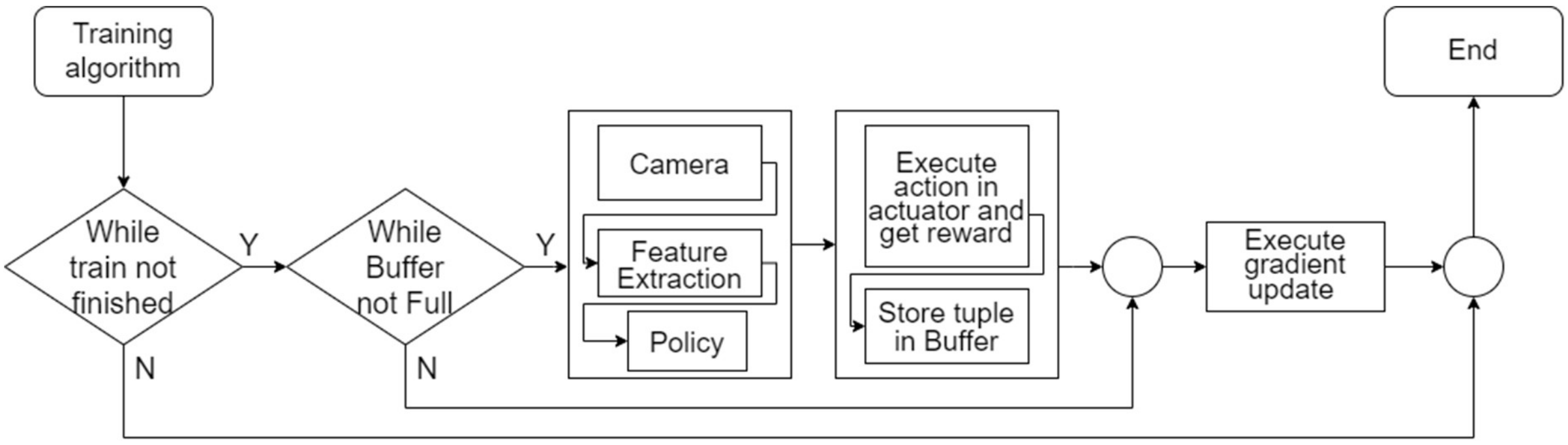

**Figure 4:** Overview of the training algorithm

## 3. Feature extraction algorithm

The feature extraction method (Figures 5 and 6) is designed to extract relevant data from the environment, focusing on the road lanes. The camera of the vehicle captures the horizon, which contains unnecessary information, as the vehicle does not require horizon data for navigation. Therefore, the first step involves cropping the camera frames to isolate the road area.

Since the road lanes are represented by white lines painted on a black surface, colour information is not essential for distinguishing the lanes. Consequently, the algorithm converts the frames to grayscale in the second step. The high resolution of the captured frames results in a larger input size for the neural network and increased memory requirements to store all the frames prior to policy training. To address this issue, the frames are downscaled, producing lower-resolution images that retain enough detail to identify the road lanes while allowing all states to be stored in a buffer before performing gradient updates.

Grayscale images contain various shades of white and black, which are irrelevant for the algorithm, as any shade of white indicates a road line. To remove this redundancy, a binary threshold filter is applied, assigning a value of zero to pixels below the threshold and a value of one to pixels above the threshold.

The final step of the feature extraction algorithm involves flattening the frame. This process removes the dimensions of the frame matrix, converting it into a one-dimensional array. This step ensures that the data is compatible with the neural network's input layer, which requires a one-dimensional input.

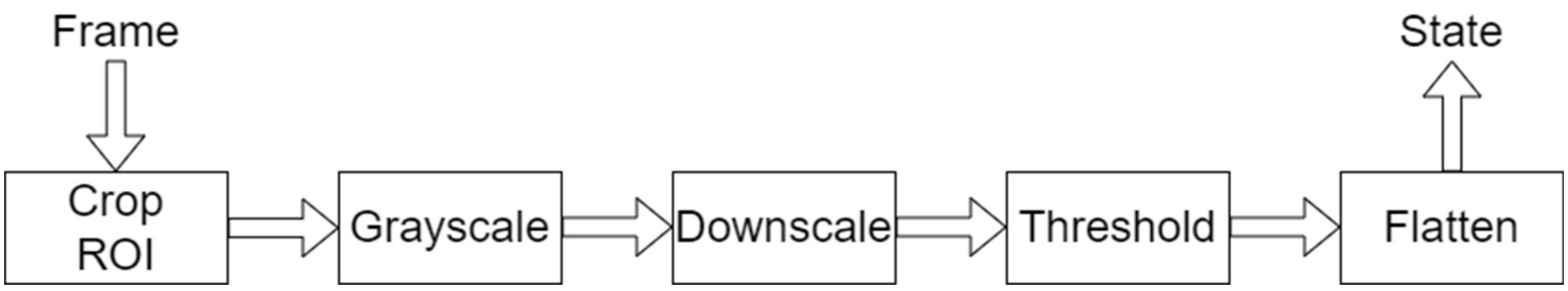

**Figure 5:** Feature extraction algorithm flowchart

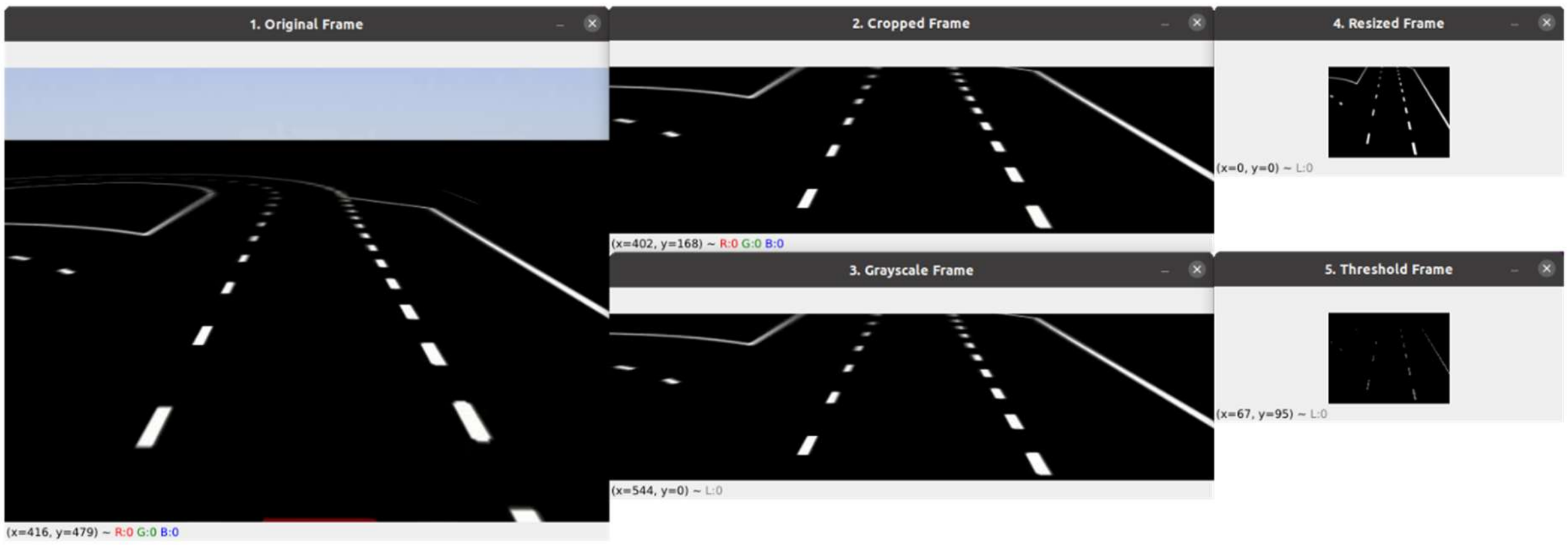

**Figure 6:** Feature extraction algorithm layers

## 4. Reward system

The reward system (Figure 8) is designed to incentivize the algorithm to develop a policy capable of effectively navigating the track. The system's inputs are the vehicle's position within the environment and the speed outputted by the policy. The vehicle's position is used to reward the algorithm for maintaining proximity to the centre of the lane. A negative reward is assigned proportional to the vehicle's deviation from the centre of the road lanes.

If the vehicle exits the track boundaries, a penalty of -10 is applied, and the episode is terminated. Detection of the vehicle's position relative to the track is determined by comparing the vehicle's displacement to a threshold value. A displacement greater than this threshold indicates that the vehicle is out of bounds.

Additionally, to ensure that the vehicle is following the correct track sequence, the sections of the track (Figure 7) are ordered. Each time the vehicle transitions to a new section, its current section is compared to the previous one. If the vehicle transitions to an incorrect section, the episode ends, and a penalty of -10 is applied.

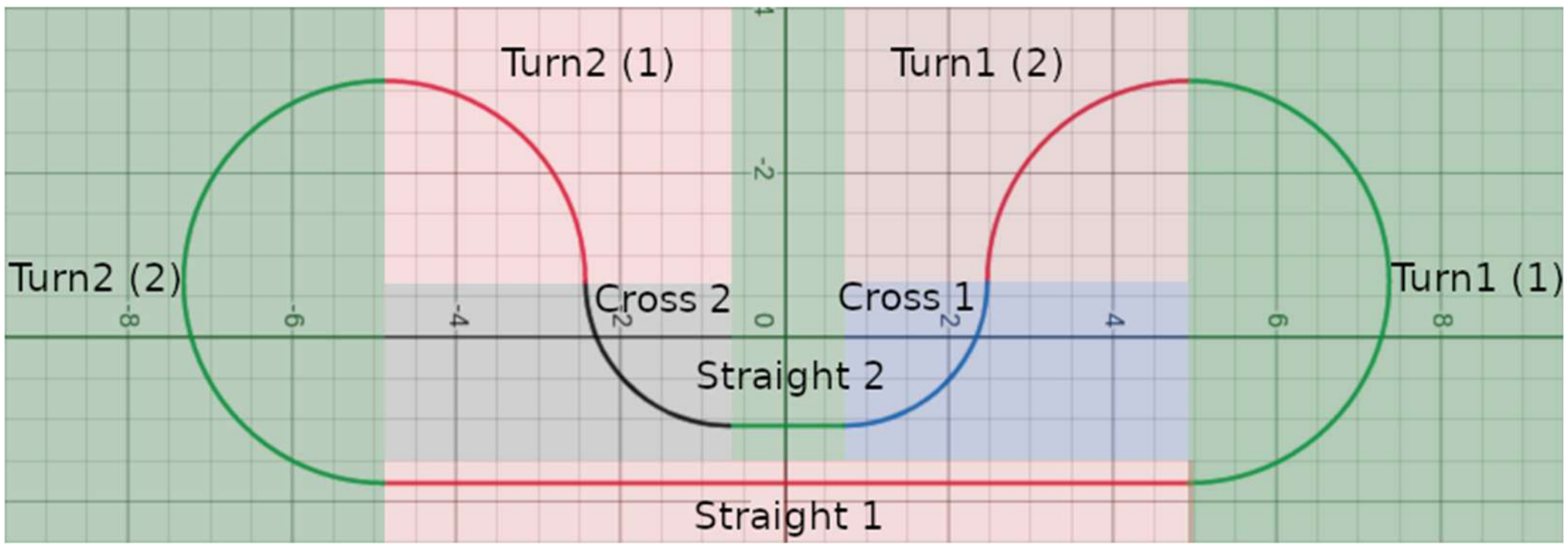

**Figure 7:** Track sections and respective ideal trajectories

The vehicle's speed is incorporated into the reward system to encourage the algorithm to develop a policy that effectively navigates the track. Initially, speed was not included in the reward structure, resulting in a policy that caused the vehicle to crash as quickly as possible, since ending the episode early was preferable to incurring penalties for failing to remain cantered in the lane. Integrating speed as a reward incentivizes the vehicle to move forward, avoiding a degenerate solution where the vehicle simply reaches the centre of the lane and remains stationary.

When the vehicle's displacement does not exceed the threshold and it follows the correct track section, the reward is calculated as the speed reward minus the vehicle's displacement from the centre of the lane, allowing the episode to continue.

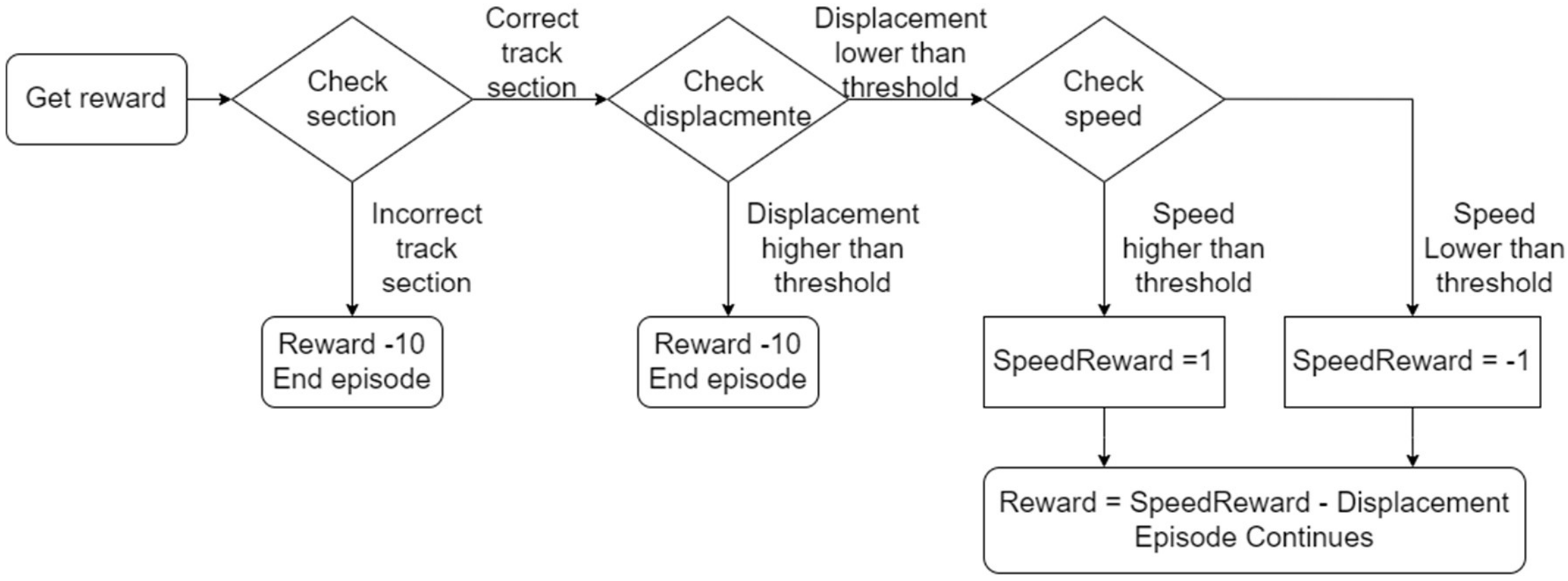

**Figure 8:** Reward system flowchart

## 5. Proximal policy optimization

### *5.1. Actor-critic*

The PPO algorithm is based on the actor-critic framework, which consists of two neural networks: the actor and the critic. The actor represents the policy and has an input layer that receives the state values, outputting an action. The critic is a neural network that also receives the state as input but outputs an estimation of the value of that state.

The value estimation provided by the critic is used to evaluate the performance of the actor. Both the actor and the critic networks are updated using gradient-based methods, improving their performance iteratively.

In the implementation of the algorithm, the loss function is defined as the mean squared error between the critic's value estimation of a state and the sum of that estimation and the advantage. The advantage represents how much better or worse the action is compared to the critic's estimation. Therefore, this formula calculates the error between the value estimation and the estimation plus the predicted error of the action's performance:

$$e = (V(s) - \hat{V}(s))^2 \qquad (1)$$

$$V(s) = \hat{A}(a, s) + \hat{V}(s) \qquad (2)$$

Although it may seem redundant to add the advantage estimation to the value estimation only to later subtract the value estimation, this step is essential. During training, the value estimation computed is only equivalent to the initial value estimation at the first gradient update. Since the value estimator undergoes multiple gradient updates, its outputs differ from the initial estimate after each update.

### *5.2. Exploration*

The training algorithm involves updating the parameters of the policy until it converges to a solution that consistently outputs the optimal action for a given state. However, if the algorithm strictly follows the policy, it will not explore alternative actions, as it lacks information about the outcomes of actions not defined by the current policy. Therefore, the algorithm requires a mechanism to explore different actions and compare their outcomes to identify the optimal action for each state.

This exploration is introduced by adding a degree of randomness to the action selection process. In the implemented algorithm (Figure 9), exploration is achieved by sampling actions from a normal distribution defined using the policy's output values. The normal distribution ensures that the actions are sampled close to those recommended by the policy, but with enough variability to enable exploration. A normal distribution is characterized by two parameters: the mean, which centres the distribution, and the deviation, which controls the degree of randomness. A higher deviation results in more exploration, while a lower deviation leads to actions that adhere more closely to the policy.

The deviation parameter is incorporated into the gradient updates, allowing the algorithm to adjust the level of exploration during training. For example, at the beginning of training, the policy may perform poorly, making random actions more beneficial than strictly following the initial, suboptimal policy. Consequently, the gradient maintains a high deviation value, promoting exploration. As training progresses and the policy improves, the deviation is reduced, enabling the algorithm to follow the increasingly optimal policy more closely. Ultimately, when the policy converges to an ideal solution, the deviation becomes minimal, and the actions follow the policy with high precision.

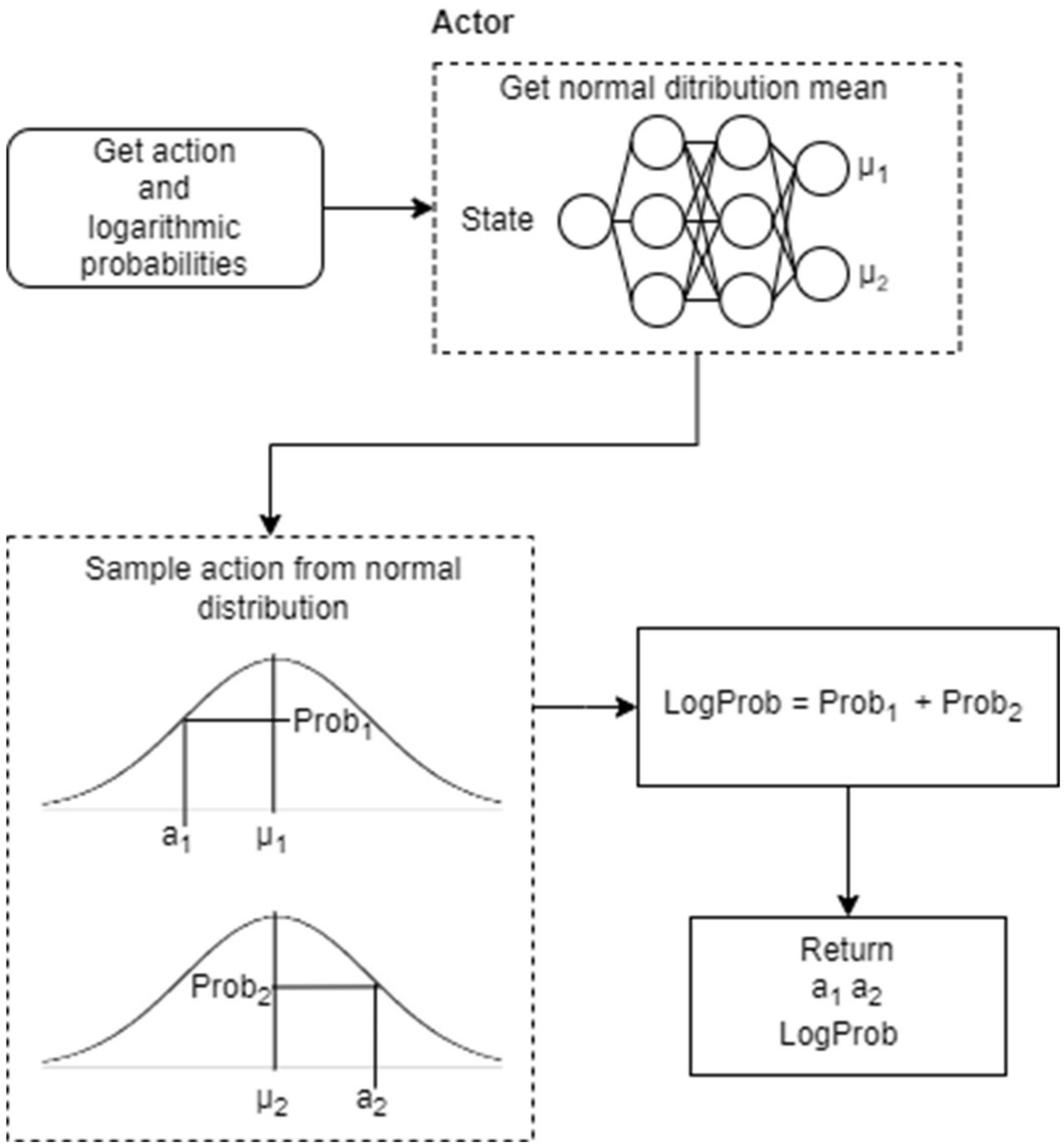

**Figure 9:** Exploration algorithm

## *5.3. Policy actions*

The outputs of the policy neural network are normalized using a sigmoid activation function, constraining the output values to lie between zero and one. The speed of the rear wheels ranges from 0 rad/s to 6 rad/s, while the front axle angle ranges from -0.27 rad to 0.27 rad. To align the neural network's outputs with the physical limits of the actuators, the two output values from the neural network are transformed using Equations (3) and (4). In these equations, $a$ represents the action value applied to the vehicle's actuators, and NN denotes the output from the neural network for both speed and angle. Figure 10 illustrates the activation function and the transformed outputs after applying Equations (3) and (4).

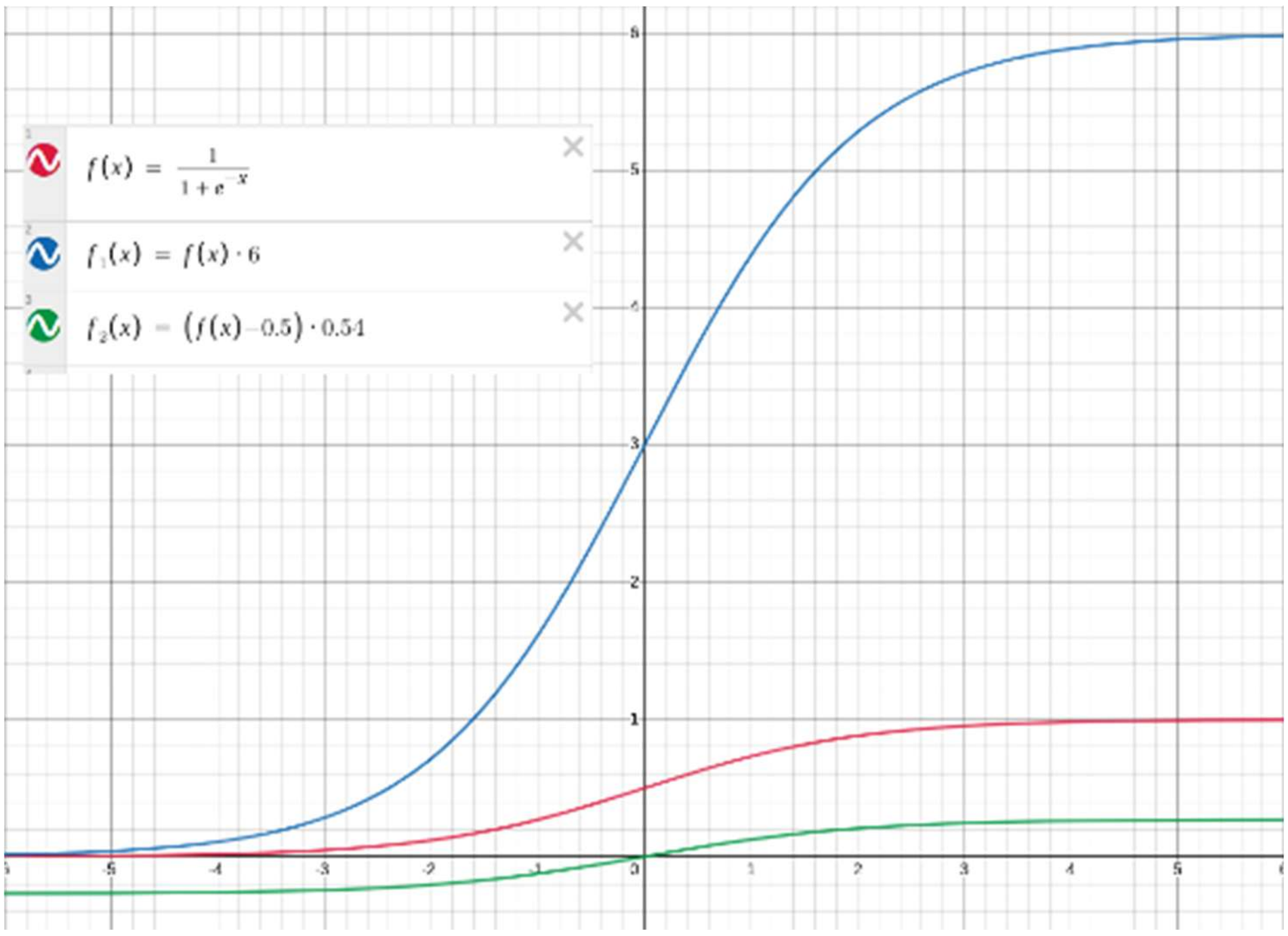

**Figure 10:** Output range and activation function

$$a_{speed} = NN_{Speed} \times 6 \qquad (3)$$

$$a_{angle} = (NN_{angle} - 0.5) \times 0{,}54 \qquad (4)$$

### *5.4. Advantage*

The advantage is given by the following formula (5)

$$A(s, a) = Q(s, a) - V(s) \qquad (5)$$

This represents the expected return of a state given an action, minus the overall expected return of that state. Intuitively, the advantage can be described as the difference between the value of a state when an action is taken and the critic's predicted value for that state.

The value of a state $V(s)$ in an actor-critic algorithm, is given by the critic estimation, so it can be represented as $\hat{V}(s)$. The hat represents the estimation. The result of this operation is also an estimation, so the advantage is represented with $\hat{A}(s, a)$. The $Q$ value represents the expected return given a specific state $s$ and action $a$ so it can be denoted as $E_\tau[G(\tau)|s_0 = s, a_0 = a]$ (equation 6). The return is defined as the sum of all discounted future rewards. This concept is expressed in equation (6) and can be used to reformulate equation (4), resulting in equation (7):

$$E_\tau[G(\tau)|s_0 = s, a_0 = a] = r_0 + \gamma V(s_1) \qquad (6)$$

$$\hat{A}(s, a) = r_0 + \gamma \hat{V}(s_1) - \hat{V}(s_0) \qquad (7)$$

Equation (7) calculates an estimate of the advantage; however, this formulation is highly dependent on the critic's estimation of the state value ($\hat{V}(s_1)$ $\hat{V}(s_0)$), resulting in significant bias. To address the issue of the advantage estimate's strong dependence on the value estimator, the temporal difference in equation (6) can be extended to any n, allowing for the derivation of an alternative advantage estimate:

$$\hat{A}^{(n)}(s,a) = r_0 + \gamma r_1 + (\dots) + \gamma^{n-1} r_{n-1} + \gamma^n \hat{V}(s_n) - \hat{V}(s_0) \qquad (8)$$

This formulation reduces the dependence of the advantage estimate on the value estimator; however, this comes at a cost, as introducing additional variables increases the variance in the advantage estimate. To address this bias–variance trade-off, a method known as Generalized Advantage Estimation was developed (Schulman et al., 2015). This technique computes an exponential average of the advantage estimates across all n values, as presented in Equations (9) and (10).

$$\hat{A}^{GAE(\gamma,\lambda)} = (1-\lambda)(\hat{A}^{(1)} + \lambda \hat{A}^{(2)} + \lambda^2 \hat{A}^{(3)} + (\dots)) \quad (9)$$

$$\hat{A}^{GAE(\gamma,\lambda)} = \sum_{l=0}^{\infty} (\lambda\gamma)^l \, \delta_{t+l}^V \qquad (10)$$

$$\delta_t^V = r_t + \gamma \hat{V}(s_{t+1}) - \hat{V}(s_t) \qquad (11)$$

In this work, Generalized Advantage Estimation is employed to compute the advantage estimate. The process involves running the algorithm in the environment and storing the rewards and states. Once the buffer containing this data is full, the critic estimates the value of each state. Using the stored rewards and the value estimates, the advantage is then calculated according to Equations (10) and (11).

### *5.5. Loss Function*

The loss function of the Proximal Policy Optimization (PPO) algorithm (Schulman et al., 2017) builds upon the Trust Region Policy Optimization (TRPO) algorithm (Schulman et al., 2015), which introduced a loss function based on trust region intervals. The TRPO loss function imposes a Kullback-Leibler (KL) divergence constraint to address issues encountered in policy gradient (PG) methods, namely the risk of making unconstrained gradient updates. Without a constraint on the gradient update step, unreliable data from the environment can lead to large, potentially destructive updates to the policy. TRPO mitigates this problem by enforcing KL divergence limits that control the step size of gradient updates.

However, calculating the KL divergence constraints precisely can be computationally challenging. PPO addresses this limitation by introducing a surrogate loss function that approximates the trust region concept while maintaining computational efficiency.

$$L^{CLIP} = E_t[\min(r_t(\theta)\hat{A}_t, clip(r_t(\theta), 1-\epsilon, 1+\epsilon)\hat{A}_t)] \quad (12)$$

$$r_t(\theta) = \frac{\pi_\theta(a_t|s_t)}{\pi_{\theta old}(a_t|s_t)} \qquad (13)$$

In the formula, $\pi_\theta(a_t|s_t)$ represents the probability of selecting action $a_t$ given state $s_t$. During training, actions are defined by sampling values from a normal distribution. The value of $\pi_\theta(a_t|s_t)$ corresponds to the probability of selecting action $a_t$ from this distribution, which is depicted in Figure 9 as Prob1 and Prob2. In the context of this project, the algorithm outputs two actions, one for speed and one for angle. Since both actions are sampled from independent normal distributions, so $\pi_\theta(a_t|s_t)$ is given by the product of the probabilities of the speed and angle actions.

To optimize computational resources, all probability operations are performed using logarithmic probabilities. By working with logarithmic values, multiplication of probabilities is replaced by addition (Figure 11), which is computationally faster. Logarithmic probabilities also simplify the calculation of ratios.

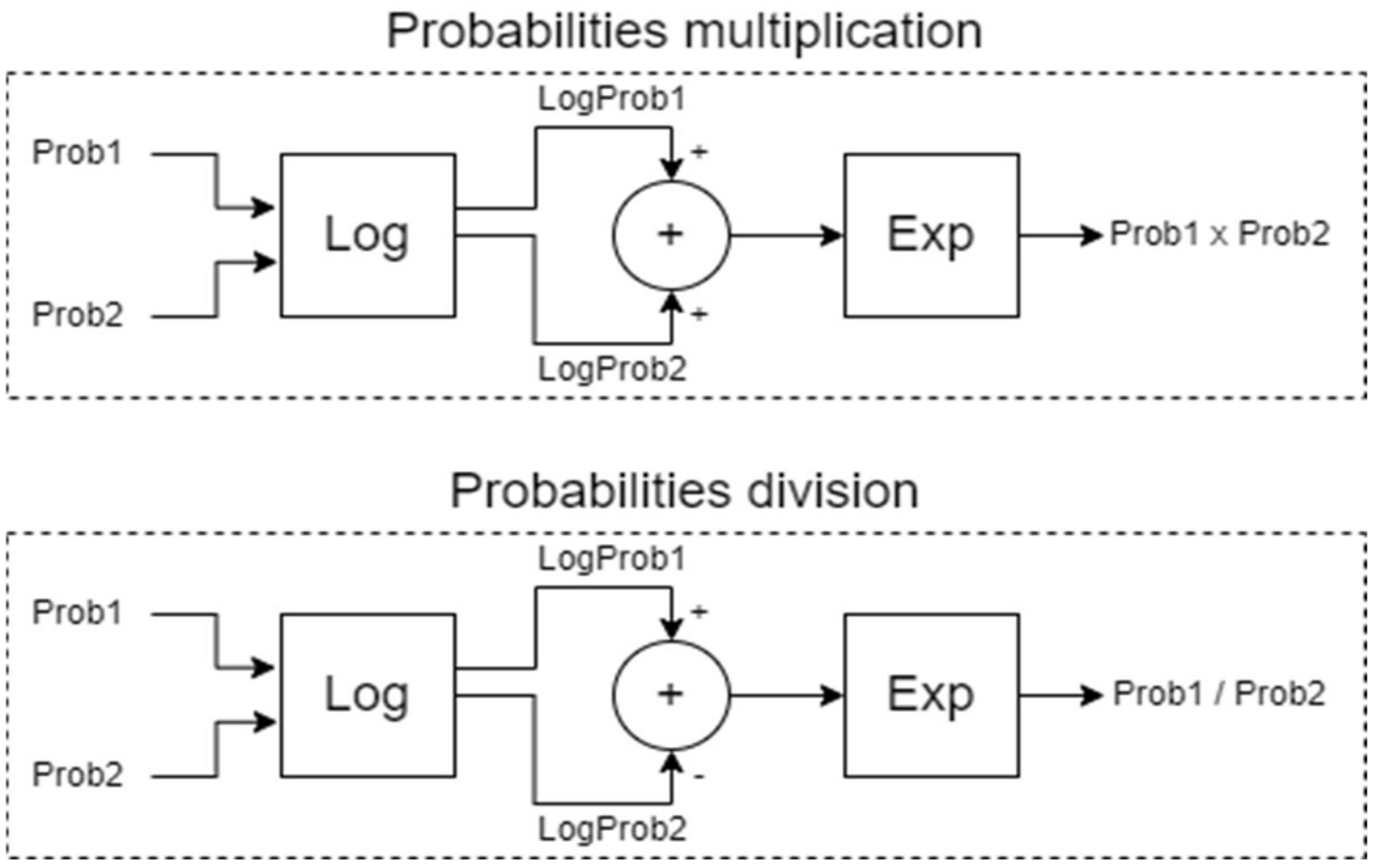

**Figure 11:** Logarithmic probabilities

The following explanation provides a clear understanding of the $L^{CLIP}$ loss function, which consists of two components: the clip function and the min function. The clip function constrains the first parameter, such that if $r_t(\theta)$ is less than $1-\epsilon$, the function returns $1-\epsilon$. Similarly, if $r_t(\theta)$ exceeds $1+\epsilon$, the function returns $1+\epsilon$.

The min function selects the smaller value between $r_t(\theta)\hat{A}_t$ and $clip(r_t(\theta), 1-\epsilon, 1+\epsilon)\hat{A}_t)$. Thus, if the advantage is positive and $r_t(\theta)$ is greater than $1+\epsilon$, the function returns $(1+\epsilon)\hat{A}_t$. If the advantage is negative and $r_t(\theta)$ is less than $1-\epsilon$ the function returns $(1-\epsilon)\hat{A}_t$. When the ration is within its limits, the value returned is $r_t(\theta)\hat{A}_t$.

In both cases, the $L^{CLIP}$ loss function imposes an upper limit on the loss value, but no lower limit, relative to the parameter change $r_t(\theta)$. This means that the algorithm is encouraged to making conservative parameter changes (Figure 12). When the ratio is high, indicating a significant parameter change, the $L^{CLIP}$ function limits this change.

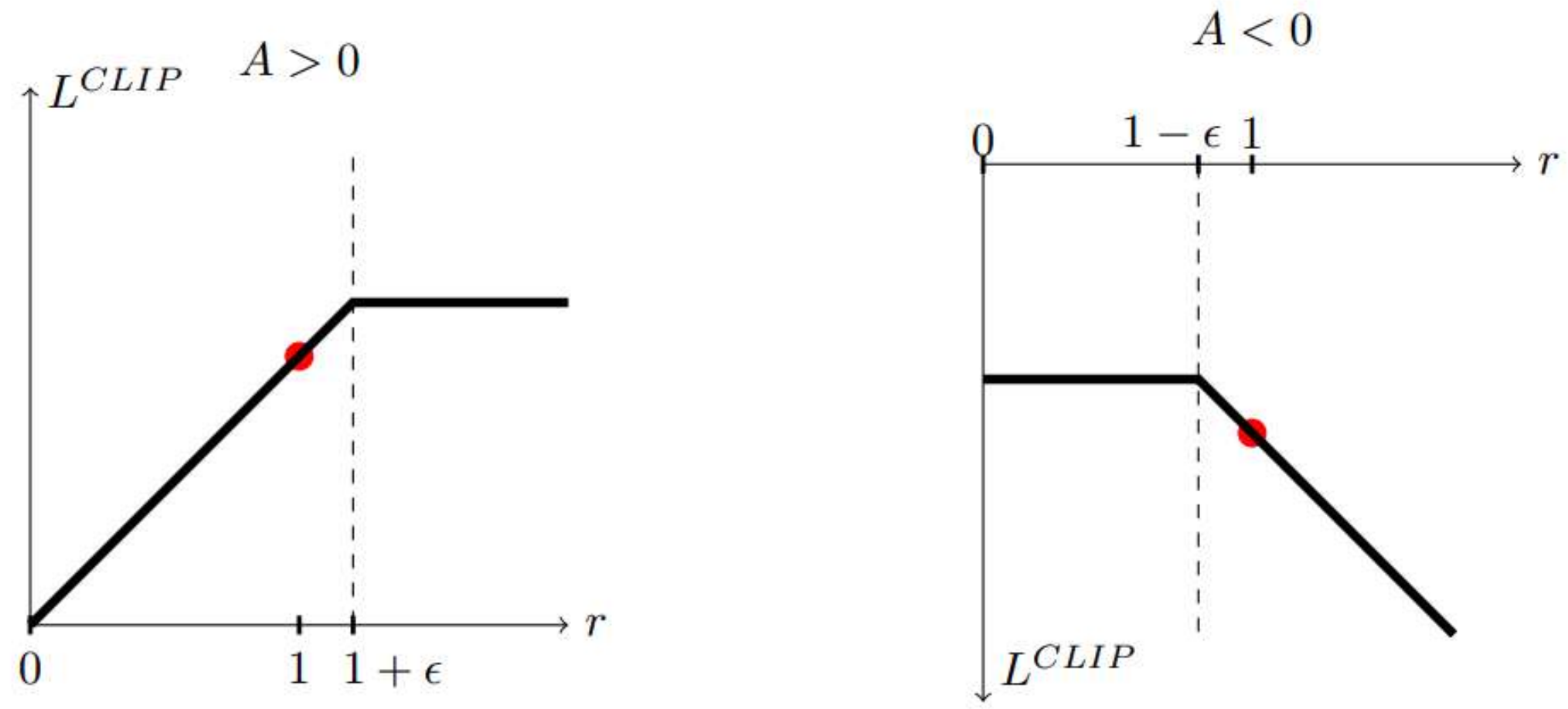

**Figure 12:** $L^{CLIP}$ chart

$$L^{CLIP+S} = E[L^{CLIP} + c_s S[\pi_\theta](s)] \qquad (14)$$

The total loss is given by the sum of $L^{CLIP}$ and the entropy term, $S[\pi_\theta](s)$, scaled by the entropy coefficient. The entropy coefficient is a constant value defined by the user. The entropy is computed using the normal distribution defined by the actor's output parameters and is calculated according to Shannon entropy.

*5.6. Mini-batch*

The algorithm collects tuples of data at each iteration. After sufficient experience has been gathered, the algorithm processes the retrieved data and assembles it into a batch containing all the information required for gradient updates. Computing the gradient using the entire batch at once is computationally intensive. To address this, the algorithm divides the batch into mini-batches and performs successive gradient updates using these smaller subsets.

However, using mini-batches can introduce a problem: gradient updates may be based on a limited and potentially unrepresentative subset of the experience. In such cases, the gradient may use insufficient data to update the policy reliably, leading to suboptimal updates. To mitigate this issue, the batch samples are randomized before being divided into mini-batches. This randomization ensures that each mini-batch is representative of the entire batch, allowing the gradient updates to reflect a comprehensive view of the experience gathered.

### *5.7. Algorithm implementation*

The implementation of the algorithm is depicted in Figure 13. It begins with the execution of the memory collector process, which runs the agent in the environment and stores, for each iteration, the current state, the action returned by the policy for that state, the reward obtained after executing the action, the next state, the 'done' flag indicating whether the episode has ended, and the probabilities of selecting the executed action.

Using the buffer returned by the memory collector process, the critic predicts the value for each stored state. These predictions, along with the rewards and the 'done' flags, are then used to compute the advantage estimates. For the number of epochs defined by the user, the batch is randomized. For each mini-batch, the gradient update is performed.

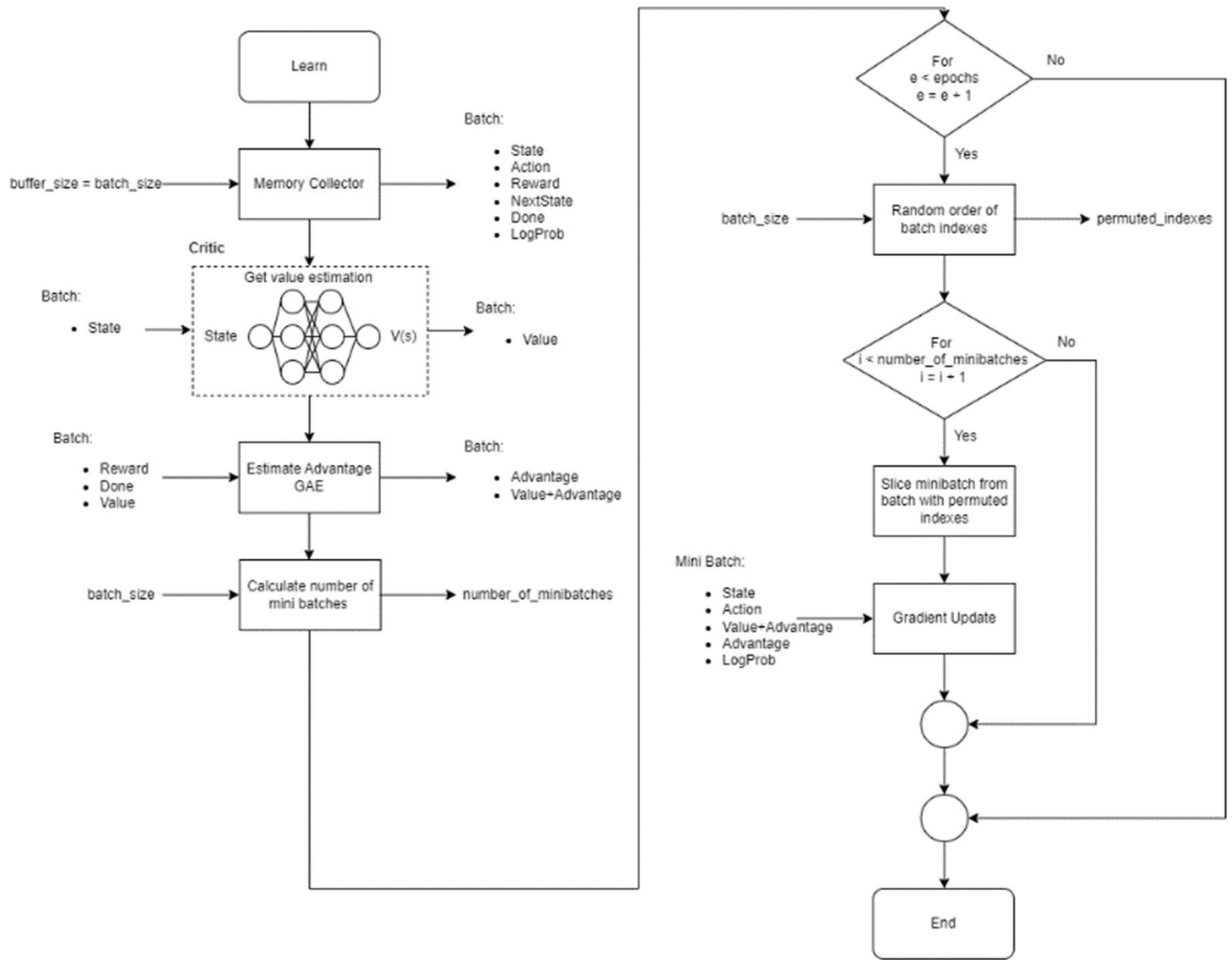

**Figure 13:** Training algorithm implementation flowchart

## 6. Results

To evaluate the performance of the algorithm in developing a policy capable of controlling the vehicle to navigate the track, two experiments were conducted. The primary metric used to compare the performance of the algorithm was the average reward achieved during each training process. Both training runs utilized the same hyperparameters, as summarized in Table I.

| | |
|---|---|
| Learning rate | $10^{-5}$ |
| Gamma | 0,99 |
| Tau | 0,95 |
| Clip | 0,2 |
| Buffer size | 4096 |
| Mini batch size | 128 |
| Epochs | 10 |

**Table I** – Hyperparameters table

### *6.1. Experiment 1*

In Experiment 1, the track is simplified and does not include parking spaces or pedestrian crossings, as illustrated in Figure 14. The vehicle begins each episode from one of the four starting points shown in Figure 14. The initial policy parameters used in this experiment were obtained from prior training, which explains why the average reward starts at a positive value. The training process is divided into two stages: the second stage (Figure 16) begins with the policy parameters learned during the first stage (Figure 15).

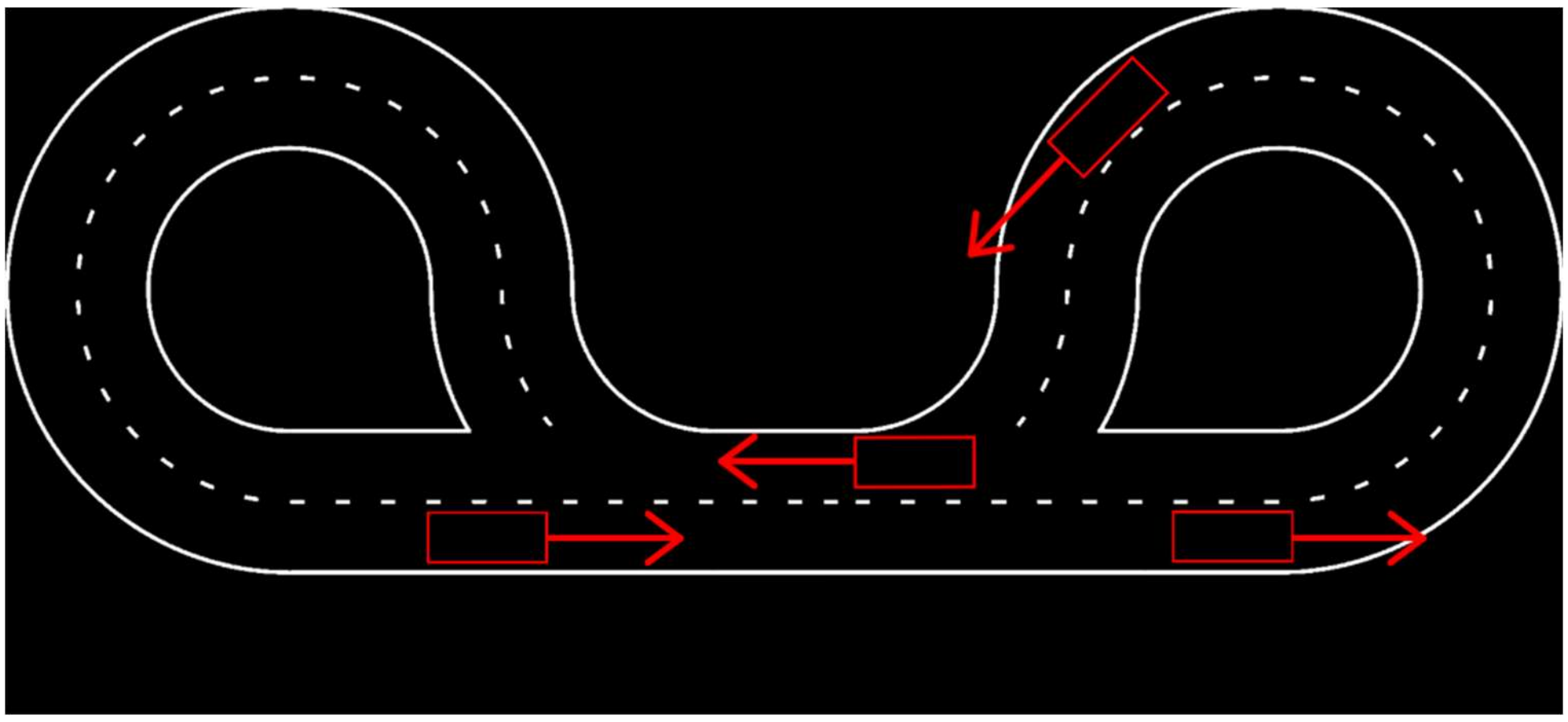

**Figure 14:** Simplified track

As shown in Figure 16, the algorithm successfully converged to a policy capable of navigating the track. This was confirmed by deploying the policy using the algorithm without the learning mechanisms, as depicted in Figure 3. During testing, the policy consistently navigated the track without crashing over five episodes of 2048 iterations each.

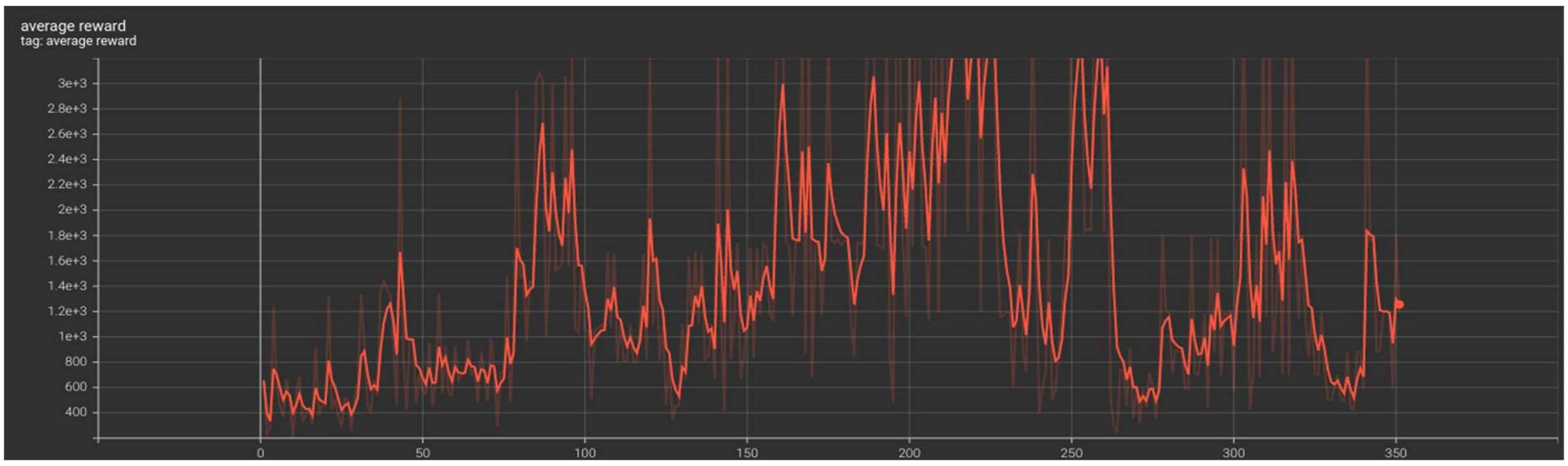

**Figure 15:** Experiment 1 training 1

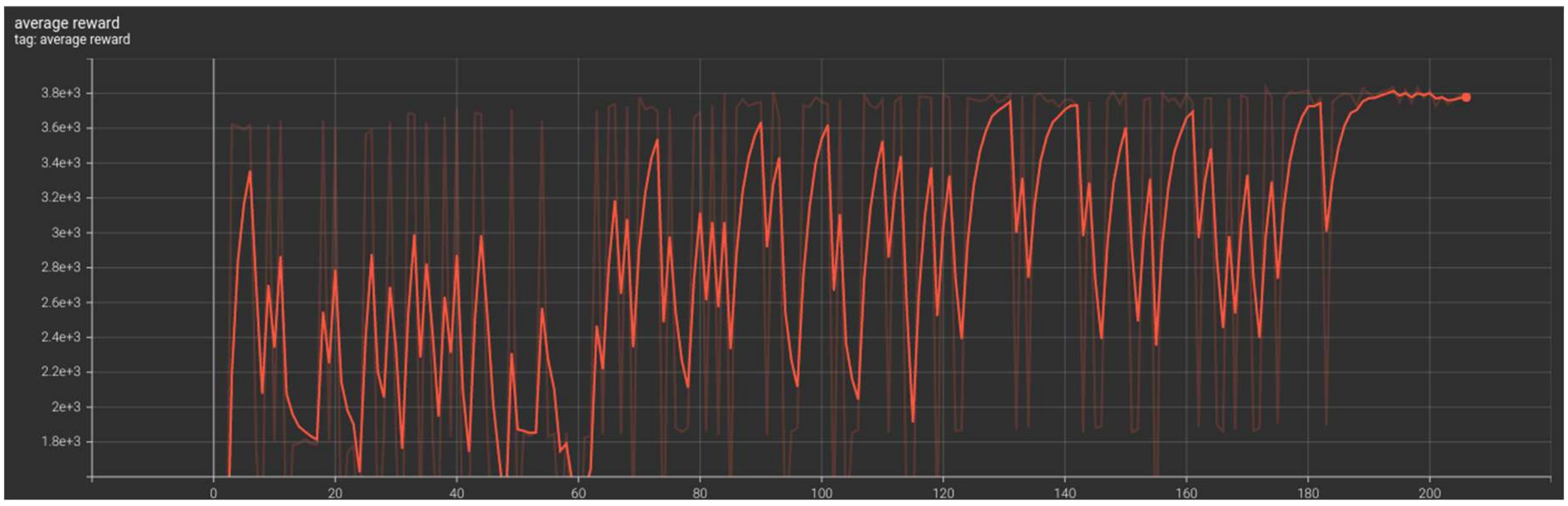

**Figure 16:** Experiment 1 training 2

## *6.2. Experiment 2*

In experiment 2 the track is the original (Figure 17), meaning that it has the pedestrians crossing and the parking spots. The vehicle starts each episode in one of the four different spots as shown in Figure 18. The starting policy parameters used in this experiment are the result of experiment 1. The algorithm failed to converge in this environment, and the training was eventually cancelled by the user.

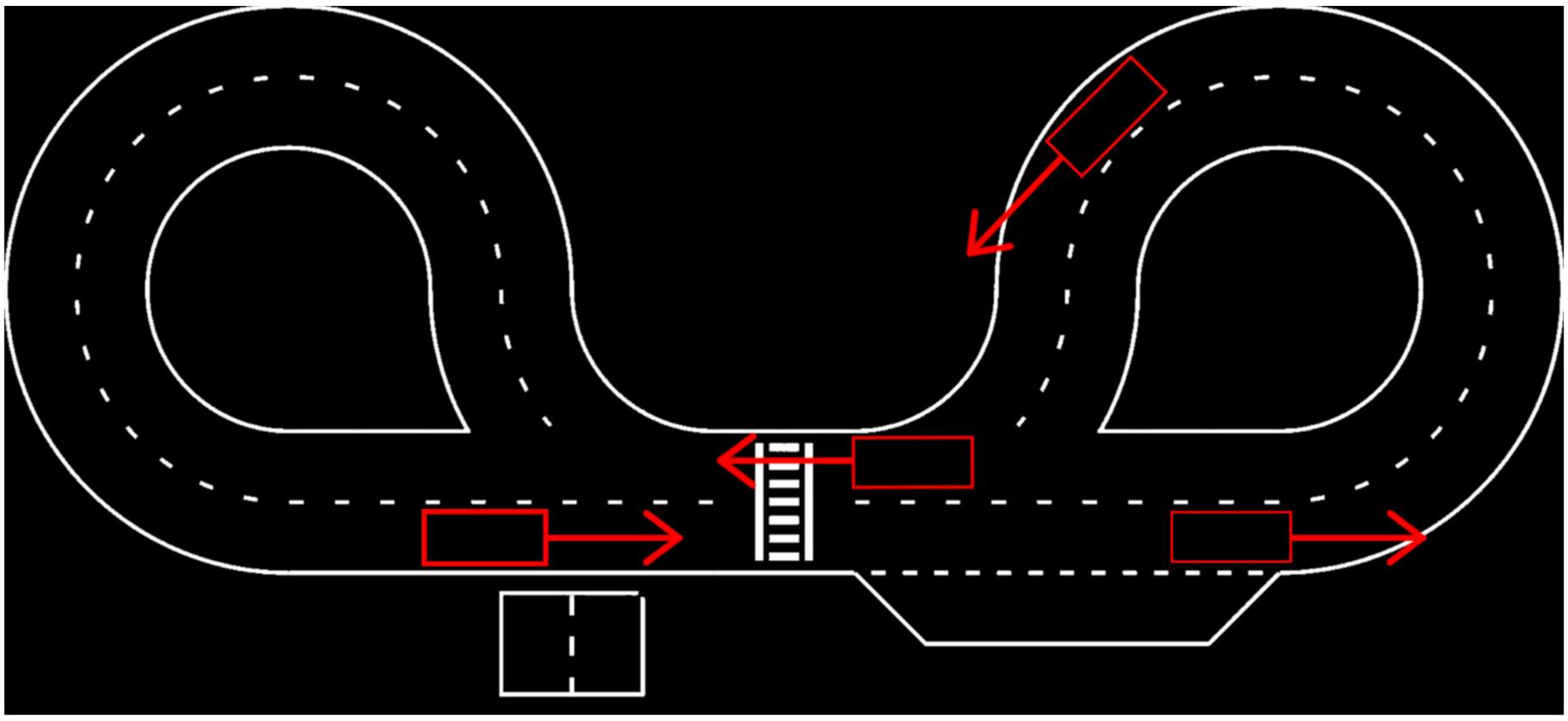

**Figure 17:** Original track

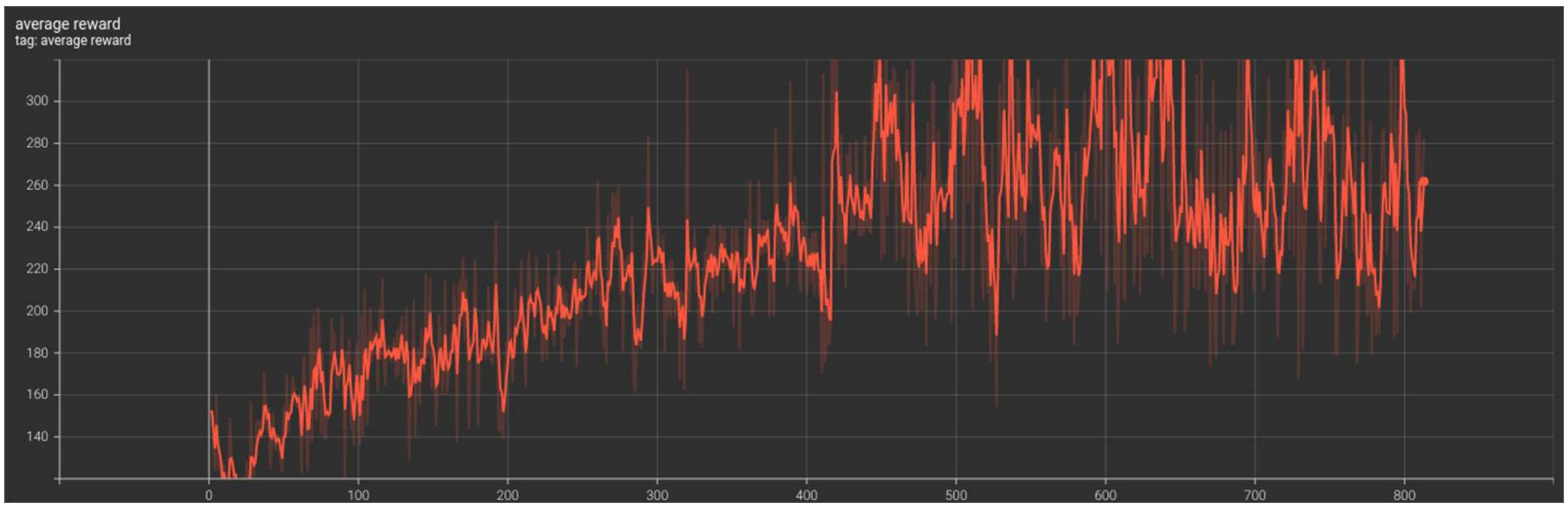

**Figure 18:** Experiment 2 training

In Experiment 1, the algorithm's state includes only the relevant data—namely, the road lines—while in Experiment 2, the state also contains noisy data, including parking spaces and pedestrian crossings. The results from these experiments demonstrate that the algorithm can learn a policy capable of navigating the track effectively when the state contains only the essential information required for navigation (road lines). However, to enable the algorithm to navigate a track that includes pedestrian crossings and parking spaces, a more robust feature extraction algorithm is needed. Such an algorithm must filter out lines representing pedestrian crossings and parking spots, retaining only the lines necessary for vehicle navigation within the road lanes. This can be achieved by implementing a line segmentation algorithm that isolates the relevant lines for navigation. Integrating this segmentation step into the feature extraction process would produce a frame containing only the road lane lines, effectively replicating the successful conditions of Experiment 1.

## 7. Conclusions

This study evaluated the use of the Proximal Policy Optimization (PPO) reinforcement learning algorithm to train a policy for controlling an autonomous vehicle model in a simulated environment. The vehicle navigates based on camera input processed through a feature extraction algorithm, with actions determined by the policy.

The experiments demonstrated that the PPO algorithm can successfully learn to navigate a simplified track when the state representation includes only the essential information, namely the road lines. However, the presence of additional visual noise, such as parking spots and pedestrian crossings, significantly affected the algorithm's performance. This highlights the importance of a robust feature extraction module capable of filtering out irrelevant information and focusing on the key elements required for navigation.

Although the training process did not reach convergence in the more complex scenario, the results from the simpler environment show promise for using reinforcement learning in autonomous driving applications. Future work should focus on enhancing the feature extraction module to handle more complex and realistic environments, as well as investigating the transferability of the trained policy to real-world implementations.

Overall, this research contributes to the development of autonomous vehicle navigation systems by demonstrating the potential of reinforcement learning algorithms in controlling vehicles based on visual input.

## 8. Further Work

Although the current implementation demonstrates the potential of reinforcement learning for autonomous vehicle navigation in controlled environments, several avenues remain open for future research and improvement:

- Sim-to-Real Transfer: Investigate the transferability of the trained policy from simulation to real-world scenarios, addressing challenges such as sensor noise, lighting variability, and hardware constraints.
- Feature Extraction Enhancement: Develop more sophisticated feature extraction techniques (e.g., convolutional neural networks) to improve robustness in complex environments, including pedestrian crossings and parking lots.

- Algorithm Comparison: Compare PPO with alternative reinforcement learning algorithms such as Soft Actor-Critic (SAC), Deep Deterministic Policy Gradient (DDPG), and Asynchronous Advantage Actor-Critic (A3C) to evaluate their performance in terms of training time, convergence, and robustness.
- Multi-Task Learning: Extend the system to handle multiple tasks simultaneously, such as obstacle avoidance and dynamic path planning, by integrating hierarchical reinforcement learning frameworks.
- Reward Shaping and Curriculum Learning: Explore reward shaping and curriculum learning strategies to accelerate the training process and improve policy generalization.

- Experimental Validation in Real Environments: Validate the proposed approach on a physical vehicle platform to assess real-world performance and to identify any discrepancies between simulation and actual deployment.
- Explainability and Safety: Investigate methods to interpret the decision-making process of the learned policy, with a focus on safety-critical applications in autonomous driving.

## Acknowledgements

The authors would like to acknowledge the Laboratory of Automation and Robotics (University of Minho, Guimarães) for their support, equipment, and valuable staff assistance.

This work has been supported by FCT – Fundação para a Ciência e Tecnologia within the R&D Units Project Scope: UID/00319/2025.

## 9. Supplementary Materials

All project files can be downloaded at the following links:

https://github.com/DinisSanchesFernandes/autonomous_vehicle

https://github.com/DinisSanchesFernandes/track_plane.

A video demonstration is available at:

https://www.youtube.com/watch?v=kLgF_g9P0iw.